\documentclass[utf8]{frontiersSCNS} 

\usepackage{url,lineno,microtype,subcaption,multirow,adjustbox,array,pifont}
\usepackage[onehalfspacing]{setspace}
\graphicspath{{img/}}

\newcolumntype{R}[2]{%
    >{\adjustbox{angle=#1,lap=\width-(#2)}\bgroup}%
    c%
    <{\egroup}%
}
\newcommand*\rot{\multicolumn{1}{R{90}{1em}}}
\newcommand*{\twoelementtable}[3][l]%
{%
    \begin{tabular}[t]{@{}#1@{}}%
        #2\tabularnewline
        #3%
    \end{tabular}%
}
\newcommand*\OK{\ding{51}}

\def\keyFont{\fontsize{8}{11}\helveticabold }
\def\firstAuthorLast{St{\"u}ber {et~al.}} 
\def\Authors{Jochen St{\"u}ber$^{1}$, Claudio Zito$^{2}$* and Rustam Stolkin$^{2}$}
\def\Address{$^{1}$IRLab, School of Computer Science, University of Birmingham, Birmingham, UK\\
$^{2}$Extreme Robotics Lab (ERL), University of Birmingham, Birmingham, UK}
\def\corrAuthor{Claudio Zito, ERL, Edgbaston, Birmingham, B15 2TT UK}

\def\corrEmail{C.Zito@bham.ac.uk}

\begin{document}
\onecolumn
\firstpage{1}

\title[Let's Push Things Forward]{Let's Push Things Forward: A Survey on Robot Pushing} 

\author[\firstAuthorLast ]{\Authors} 
\address{} 
\correspondance{} 

\extraAuth{}

\maketitle

\begin{abstract}

\section{As robot make their way out of factories into human environments, outer space, and beyond, they require the skill to manipulate their environment in multifarious, unforeseeable circumstances. With this regard, pushing is an essential motion primitive that dramatically extends a robot's manipulation repertoire. In this work, we review the robotic pushing literature. While focusing on work concerned with predicting the motion of pushed objects, we also cover relevant applications of pushing for planning and control. Beginning with analytical approaches, under which we also subsume physics engines, we then proceed to discuss work on learning models from data. In doing so, we dedicate a separate section to deep learning approaches which have seen a recent upsurge in the literature. Concluding remarks and further research perspectives are given at the end of the paper.}

\tiny
 \keyFont{ \section{Keywords:} robotics, pushing, manipulation, forward models, motion prediction } 
\end{abstract}

\section{Introduction}

We argue that pushing is an essential motion primitive in a robot's manipulative repertoire. Consider, for instance, a household robot reaching for a bottle of milk located in the back of the fridge. Instead of picking up every yoghurt, egg carton, or jam jar obstructing the path to create space, the robot can use gentle pushes to create a corridor to its  lactic target. Moving larger obstacles out of the way is even more important to mobile robots in environments as extreme as abandoned mines \citep{thrun2004mines}, the moon \citep{king2016nonprehensile}, or for rescue missions as for the Fukushima Daiichi Nuclear Power Plant. In order to save cost, space, or reduce payload, such robots are often not equipped with grippers, meaning that prehensile manipulation is not an option. Even in the presence of grippers, objects may be too large or too heavy to grasp. 

In addition to the considered scenarios, pushing has numerous beneficial applications that come to mind less easily. For instance, pushing is effective at manipulating objects under uncertainty \citep{brost1988uncertaingrasp,dogar2010pushgrasp}, and for pre-grasp manipulation, allowing robots to bring objects into configurations where they can be easily grasped \citep{king2013pregrasp}. Less existential, yet highly interesting and entertaining, dexterous pushing skills are also widely applied and applauded in robot soccer \citep{emery2001nonholocontrol}. 

Humans perform skilful manipulation tasks from an early age on, and are able to transfer behaviours learned on one object to objects of novel sizes, shapes, and physical properties. For robots, achieving those goals is challenging. For one thing, this complexity arises from the fact that frictional forces are usually unknown but play a significant role for pushing \citep{zhou2016polynomial}. Furthermore, the dynamics of pushing are highly non-linear, with literal tipping points, and sensitive to initial conditions \citep{yu2016data}. The large body of work on robotic pushing has nevertheless produced many accurate models for predicting the outcome of a push, some analytical, some data-driven. However, models that generalise to novel objects are scarce \citep{kopicki2016forwardmodel,stuber2018icra}, highlighting the demanding nature of the problem.

In this paper, we review the robotic pushing literature. We focus on work concerned with making predictions of the motion of pushed objects, but we also cover relevant applications of pushing for planning and control. We begin with analytical approaches, under which we also subsume physics engines, we then proceed to discuss data-driven approaches as well as deep learning approaches which have recently become very popular in the literature.

\section{Problem Statement}

\begin{figure}[t]
    \centering
    \includegraphics[scale=0.5]{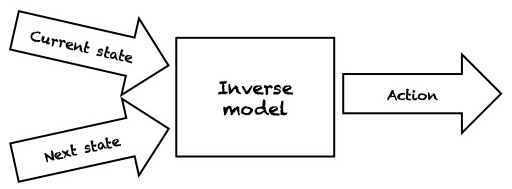}

    \caption{An inverse model computes an action which will affect the environment such that the next desired state (or configuration) is achieved from the current state.}
    \label{fig:inv_model}
\end{figure}

\begin{figure}[t]
    \centering
    \includegraphics[scale=0.5]{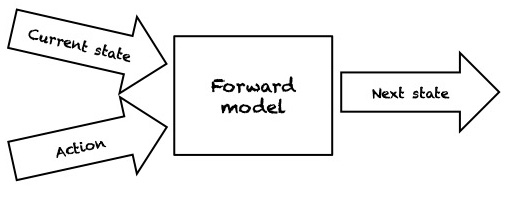}
    \caption{A forward model makes a prediction on how an action will affect the current state of the environment by returning the configuration after the action is taken.}
    \label{fig:for_model}
\end{figure}

Even in ideal conditions, such as \emph{structured environments} where an agent has a complete model of the environment and perfect sensing abilities, the problems of robotic grasping and manipulation are not trivial. By complete model of the environment we mean that physical and geometric properties of the world, such as pose, shape, friction parameters and the mass of the object we wish to manipulate, are exactly known.  In fact, the object to be manipulated is indirectly controlled by contacts with a robot manipulator (e.g. pushing by a contacting finger part), and an \emph{inverse model} (IM), which computes an action to produce the desired motion or set of forces on the object, may not be known. Sometimes \emph{forward models} (FM) may be fully or partially known, even where IMs are not available. In such cases, an FM can be used to estimate the next state of a system, given the current state and a set of executable actions. This enables planning to be achieved by imagining the likely outcomes from all possible manipulative actions, and then choosing the action which achieves the most desirable end state. Figures~\ref{fig:inv_model} and ~\ref{fig:for_model} show a graphical representation of these two models.
However the manipulation and grasping problem is typically  defined in continuous state and action spaces, hence it is computationally intractable to build an optimal sequence of actions, or plan, by exploring all possible action-state combinations.

Even more challenging is the problem of grasping and manipulation in \emph{unstructured environments}, where these ideal conditions do not exist. There are several reasons why an agent may fail to build a complete description of the state of the environment: sensors are noisy, robots are difficult to calibrate, actions' outcomes are unreliable due to unmodelled variables (e.g. friction, mass distribution). Uncertainty can be modelled in several ways, but in the case of manipulation there are typically two types of uncertainty:
\begin{itemize}
\item \emph{Uncertainty in physical effects}: occurs when the robot acts on external bodies via physical actions (e.g., contact operations). This interaction transforms the current state of the world according to physical laws which are not fully predictable. For example, a pushed object may slide, rotate or topple with complex motions which are extremely difficult to predict, and involve physical parameters which may not be known. We can think of this as uncertainty on future states.

\item \emph{Uncertainty in sensory information}: occurs when some of the quantities that define the current state of the world are not directly accessible to the robot. Thus the necessity to develop strategies to allow the robot to complete tasks in partial ignorance by recovering knowledge of its environment. In such cases, there is uncertainty about how much new information will be yielded during the execution of a new robotic action.
\end{itemize} 

This paper is concerned with the evolution of FMs and their application in robotics. Table~\ref{tab:at_glance} summarises the literature at glance. 
The papers are classified according to the type of approach implemented. We identify the following six classes.
\begin{enumerate}
    \item \textbf{Purely analytical}. It is mostly seminal work drawn from classical mechanics. 
    \item \textbf{Hybrid}. It extends analytical approaches with data-driven methods. Whilst the interactions between objects are still represented analytically, some quantities of interest are estimated based on observations, e.g. the coefficients of friction.
    \item \textbf{Dynamic analysis}. It integrates dynamics in the model. 
    \item \textbf{Physics engines}. It employs a physics engine as a ``black box'' to make predictions about the interactions.
    \item \textbf{Data-driven}. It learns how to predict physical interaction from examples.
    \item \textbf{Deep learning}. As the data-driven approaches, it learns how to construct an FM from examples. The key insight is that the deep learning approaches are based on feature extraction.
\end{enumerate}
The features highlighted for each approach are as follows.
\begin{itemize}
    \item The assumptions made by the authors on their approach. We highlight i) the quasi-static assumption in the model, ii) if it is a seminal work on 2D shapes, and iii) if the method required a known model of the object to be manipulated.
    \item The type of motion analysed in the paper, such as 1D, planar (2D translation and 1D rotation around the $x-$axis), or full 3D (3D translation and 3D rotation).
    \item The aim of the paper. We distinguish between predicting the motion of the object, estimating physical parameters, planning pushes, and analysing a push to reach a stable grasp.
    \item The model. We distinguish between analytical, constructed from data, and by using a physics simulator.
\end{itemize}

\section{Analytical Approaches}

\subsection{Quasi-Static Planar Pushing}

Early work on robotic pushing focused on the problem of quasi-static planar pushing of sliding objects. In a first phase, several researchers, following pioneering work by Matthew T. Mason, approached the problem analytically, explicitly modelling the objects involved and their physical interactions whilst drawing on theories from classical mechanics. More recently, this tradition has moved to extend analytical models with more data-driven methods.

\begin{table*}[t]
    \centering
    \footnotesize
    \begin{tabular}{|c|l|ccc|ccc|cccc|ccc|}
        \hline
        \hline
         && \multicolumn{3}{c|}{Assumptions} & \multicolumn{3}{c|}{Motion} & \multicolumn{4}{c|}{Aim} &  \multicolumn{3}{c|}{Model} \\ \hline
         && \rot{\twoelementtable{Quasi-static}{assumption}} \hspace{1mm} & \rot{2D Object} & \rot{\twoelementtable{Known}{object}} \hspace{1mm} \vline & \rot{1D} & \rot{Planar} & \rot{3D} \vline &  \rot{\twoelementtable{Motion}{ prediction}} \hspace{1mm} & \rot{\twoelementtable{Parameter}{estimation}} \hspace{1mm} & \rot{\twoelementtable{Path}{planning}} \hspace{1mm} & \rot{Grasping} \vline & \rot{Analytical} & \rot{Data-driven} & \rot{\twoelementtable{Physics}{simulator}} \hspace{1mm} \vline  \\\hline
         \multirow{22}{*}{PA} &\citet{mason1982manipulation}  
                                        &    & \OK& \OK &\OK&   &\OK&   &   &   &\OK&   &  &\\
         &\citet{mason1986quasistatic}   & \OK& \OK& \OK &   &\OK&   &\OK&   &   &   &\OK&   & \\
         &\citet{peshkin1988motionslide, peshkin1988slide} 
                                        & \OK&\OK&\OK& &    &\OK&   &\OK&   &   &   &\OK&   \\
         &\citet{goyal1991planarfirst}   & \OK&\OK&\OK &   & \OK     &&   &\OK&   &   &&&\\
         &\citet{alexander1993bounds}    & \OK& &\OK &   &\OK&   &\OK&   &   &   &\OK&   &\\
         &\citet{lee1991fixture}         & \OK&\OK&\OK&    &\OK&   &\OK&   &   &   &\OK&   &\\
         &\citet{lynch1992tactilefeedback}& \OK& &\OK&  &\OK&   &   &\OK&   &   &\OK&   &  \\
         &\citet{howe1996approxlimit}    &    & &\OK&   &\OK&   &\OK&   &   &   &\OK&   &   \\
         &\citet{mason1990blockwall}     &    &  &\OK&  &   &\OK&   &   &\OK&   &\OK&   &   \\
         &\citet{mayeda1991wallpush}     &    & &\OK&   &\OK&   &   &   &\OK&   &\OK&   &  \\
         &\citet{akella1992poseplane, akella1998posing} 
                                        &    &&\OK&    &\OK&   &   &   &\OK&   &\OK&   &   \\
         &\citet{narasimhan1994task}     &\OK & \OK&\OK&   &\OK&   &   &   &\OK&   &\OK&   &\\
         &\citet{lynch1996stablepush}    &\OK & \OK&\OK&   &\OK&   &   &   &\OK&   &\OK&   &\\
         &\citet{agarwal1997nonholonomic}&    & \OK&\OK&   &\OK&   &   &   &\OK&   &\OK&   &\\
         &\citet{nieuwenhuisen2005pushplandisk}
                                        &\OK &\OK&\OK&    &\OK&   &   &   &\OK&   &\OK&   & \\
        &\citet{deberg2010pushplandiskshaped}
                                        &    &\OK&\OK&    &\OK&   &   &   &\OK&   &\OK&   & \\
        &\citet{miyazawa2005graspless}   &    &\OK& &\OK &   &   &   &   &\OK&   &\OK&   &  \\
        &\citet{cappelleri2006micromanipulation} 
                                        & \OK&&\OK &    &\OK&   &   &   &\OK&   &\OK&   &  \\
         &\citet{dogar2011framework}     & \OK&&\OK &    &   &\OK&   &   &   &\OK&\OK&   &  \\
         &\cite{cosgun2011pushplancluttered}
                                        &    &\OK&\OK &    &\OK&   &   &   &\OK&   &   &   &\\
          &\citet{lee2015multicontact}   &    &\OK&\OK &    &\OK&   &   &   &\OK&   &   &   &\\                          
         &\citet{king2016nonprehensile}  &    &&\OK &    &\OK&   &   &   &\OK&   &   &   &\OK\\
         \hline
         \multirow{5}{*}{HD} &\citet{lynch1993estimate}      &&\OK &    &    &\OK&   &   &\OK&   &   &\OK& &   \\
         &\citet{yoshikawa1991estimate}  &    &&\OK &    &\OK&   &   &\OK&   &   &\OK&   &   \\
         &\citet{ruizugalde2010objectmodel, ruizugalde2010effectawarepushing} 
                                        &    &&\OK &    &\OK&   &   &\OK&   &   &\OK&   &   \\
         &\citet{zhu2017physicsengine}   &    &&\OK &    &   &\OK&   &   &\OK&   &   &   &   \\
         &\citet{bauza2017planarprob}    & \OK&&\OK &    &\OK&   &   &   &\OK   &   &\OK&   & \\
         \hline
         \multirow{4}{*}{DA} &\citet{brost1992dynamic}       &    && &    &\OK&   &\OK&   &   &   &\OK&   &   \\
         &\citet{jia1999pose}            &    &&\OK &    &\OK&   &\OK&   &   &   &\OK&   &  \\
         &\citet{behrens2013dynamic}     &    &&\OK &    &\OK&   &\OK&   &   &   &\OK&   &   \\
         &\cite{chavan2015inhand}        &    &&\OK &    &   &\OK&\OK&   &   &\OK&   &\OK&  \\
         \hline
         \multirow{3}{*}{PE} &\citet{zito2012two}            & \OK&&\OK &    &   &\OK&   &   &\OK &   &   &\OK& \\
         &\citet{scholz2014oomdp}        &    &&\OK &    &\OK&   &  &\OK &  &    &   &\OK&\OK\\
         &\citet{zhu2017physicsengine}   &    &&\OK &    &   &\OK&  &\OK &  &    &   &   &\OK\\
         \hline
         \multirow{9}{*}{DD} &\citet{moldovan2012learning}   &    &&\OK &    &\OK&   &\OK&   &   &   &   &\OK&  \\
         &\citet{ridge2015pushaffordances} &    && &    &   &\OK&\OK&   &   &   &   &\OK&  \\
         &\citet{zrimec1991pushlearninduction} 
                                            &    &&\OK &    &   &\OK&\OK&   &   &   &   &\OK&\\ 
         &\citet{salganicoff93avisionbased} 
                                        &    &&\OK &    &\OK&   &\OK&   &   &   &   &\OK& \\
         &\citet{walker2009maps}         &    && &    &\OK&   &\OK&   &   &   &   &\OK&\\
         &\citet{lau2011pushstrategy}    &    && &    &\OK&   &\OK&   &   &   &   &\OK& \\
         &\citet{kopicki2011predictrigid, kopicki2016forwardmodel} 
                                        & \OK&&\OK &    &   &\OK&\OK&   &   &   &   &\OK&\\
         &\cite{stuber2018icra}          & \OK&& &    &   &\OK&\OK&\OK&   &   &   &\OK&\\
         &\citet{mericli2015complexpush} & \OK&& &    &\OK&   &\OK&   &   &   &   &\OK& \\
         \hline
         \multirow{7}{*}{DL}&\citet{denil2016physicsexperiments}
                                        &    && &    &   &\OK&   &\OK&   &   &   &\OK& \\
         &\citet{chang2016deepdynamics}  &    && &    &\OK&   &   &\OK&   &   &   &\OK& \\
         &\citet{watters2017interactnetworks}
                                        &    &&\OK &    &\OK&   &\OK&   &   &   &   &\OK&  \\
         &\citet{fragkiadaki2015predictbilliard}
                                        &    &&\OK &    &\OK&   &\OK&   &   &   &   &\OK&  \\
         &\citet{ehrhardt2017deepphysics}
                                        &    &&\OK &    &\OK&   &\OK&   &   &   &   &\OK& \\
         &\citet{byravan2016deeprigid}   &    &&\OK &    &   &\OK&\OK&   &   &   &   &\OK&  \\
         &\cite{finn2016unsupervisedphysics}
                                        &    && &    &   &\OK&\OK&   &   &   &   &\OK&  \\
         
         \hline
         \hline
    \end{tabular}
    \caption{Summary of the literature at glance. PA: Purely Analytical; HD: Hybrid; DA: Dynamic Analysis; PE: Physics Engines; DD: Data Driven; DL: Deep Learning. }
    \label{tab:at_glance}
\end{table*}

\subsubsection{Purely Analytical Approaches}

To briefly introduce the problem, \textit{planar pushing} \citep{mason1982manipulation}, refers to an agent pushing an object such that pushing forces lie in the horizontal support plane while gravity acts along the vertical. Both pusher and pushed object move only in the horizontal plane, effectively reducing the world to 2D. Meanwhile, the \textit{quasi-static assumption} \citep{mason1986quasistatic} in this context means that the involved objects' velocities are small enough that inertial forces are negligible. In other words, objects only move when pushed by the robot. Instantaneous motion is then the consequence of the balance between contact forces, frictional forces, and gravity. The quasi-static assumption makes the problem more tractable, yielding simpler models. A key challenge in predicting the motion of a pushed object under manipulation is that the distribution of pressure at the contact between object and supporting surface is generally unknown. Hence, the system of frictional forces that arise at that contact is also indeterminate \citep{mason1982manipulation}. 

\begin{figure}
    \centering {
    \includegraphics[scale=0.5]{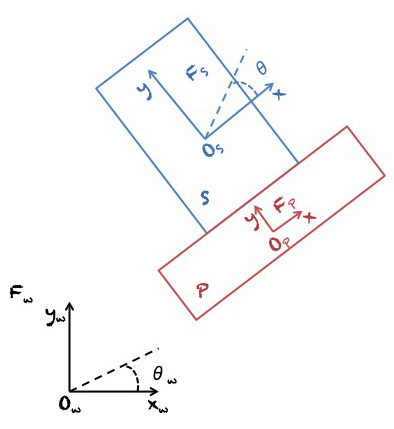}
    \includegraphics[scale=0.5]{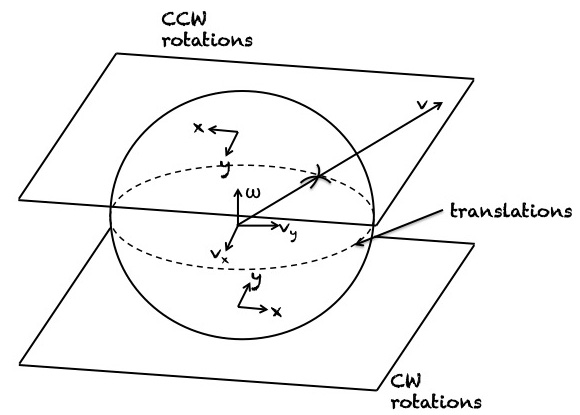}
    }
    \caption{The slider $S$ (blue) is a rigid object in the plane $\mathbb{R}^2$, and its configuration space is $\mathbb{R}^2\times \Theta$, i.e. 2D translation and one rotation over the $x-$axis. The slider is pushed by a rigid pusher $P$ (red) at a point or set of points of contact. A world frame $F_w$ with origin $O_w$ is fixed in the plane, and a slider frame $F_w$ with origin $F_s$ is attached to the centre of friction of the slider $S$. The configuration $(x_w,y_w,\theta_w)^\top$ describes the position and orientation of the slider frame $F_s$ relative to the world frame $F_w$. Similarly, a pusher frame $F_p$ with origin $O_p$ and its configuration is computed. On the right side of the figure, the relation between the unit motion vector $\mathbf{v}=\{v_x,v_y,\omega\}^\top$ and the centre of rotation of frame $F_s$ is described by the projection shown from the unit motion sphere to the tangent planes (one for each rotation sense). The line at the equator of the sphere represents translations. 
    Reproduced from~\cite{lynch1996stablepush}.}
    \label{fig:planar_pushing}
\end{figure}

\citet{mason1982manipulation, mason1986pushplan} started the line of work on pushing, proposing the voting theorem as a fundamental result. It allows to find the sense of rotation of a pushed object given the pushing direction and the centre of friction without requiring knowledge of the pressure distribution. Drawing on this seminal work, \citet{peshkin1988motionslide, peshkin1988slide} found bounds on the rotation rate of the pushed object given a single-point push. Following that, \citet{goyal1991planarfirst} introduced the \textit{limit surface} which describes the relationship between the motion of a sliding object and the associated support friction given that the support distribution  is completely specified. Under the quasi-static assumption, the limit surface allows to convert the generalised force applied by a pusher at a contact to the instantaneous generalised velocity of the pushed object. \citet{alexander1993bounds} considered the case when only the geometric extent of the support area is known, and described techniques to bound the possible motions of the pushed object. While the limit surface provides a powerful tool for determining the motion of a pushed object, there exists no convenient explicit form to construct it. In response to this challenge, \citet{lee1991fixture} proposed to approximate the limit surface as an ellipsoid to improve computational time. However, their approximation requires knowledge of the pressure distribution. Marking a milestone of planar pushing research, \citet{lynch1992tactilefeedback} applied the ellipsoidal approximation to derive a closed-form analytical solution for the kinematics of quasi-static single-point pushing, including both sticking and sliding behaviours. Subsequently, \citet{howe1996approxlimit} explored further methods for approximating limit surfaces, including guidance for selecting the appropriate approach based on the pressure distribution, computational cost, and accuracy. 

Results on the mechanics of planar pushing have been used for \textit{planning and control} of manipulator pushing operations. To begin with, \citet{mason1990blockwall} showed how to synthesize robot pushing motions to slide a block along a wall, a problem later also studied by \citet{mayeda1991wallpush}. \citet{akella1992poseplane, akella1998posing} analysed the series of pushes needed to bring a convex polygon to a desired configuration. \citet{narasimhan1994task} and \citet{kurisu1995trajectory}  studied the problem of moving an object among obstacles by pushing with point contact. \citet{lynch1996stablepush} comprehensively studied stable pushing of a planar object with a fence-shaped finger, considering mechanics, control, and planning. First, they derived conditions for stable edge pushing, considering the case where the object will remain attached to the pusher without slipping or breaking contact. Based on this result, they then used best-first search to find a path to a specified goal location. \citet{agarwal1997nonholonomic} proposed an algorithm for computing a contact-preserving push plan for a point-sized pusher and a disk-shaped object. \citet{nieuwenhuisen2005pushplandisk} utilised compliance of manipulated disk-shaped objects against walls to guide their motion. They presented an exact planning algorithm for 2D environments consisting of non-intersecting line segments. Subsequently, \citet{deberg2010pushplandiskshaped} improved this approach from a computational perspective and presented push planning methods both for the contact-preserving case and less restrictive scenarios. \citet{miyazawa2005graspless} used a rapidly-exploring tree (RRT) \citep{lavalle1998rrt} for  planning non-prehensile manipulation, including pushing, of a polyhedron with three degrees of freedom (DOF) by a robot with spherical fingers. They do not allow for sliding and rolling of robot fingers on the object surface. \citet{cappelleri2006micromanipulation} have solved a millimetre scale 2D version of the peg in the hole problem, using Mason's models for quasi-static manipulation and an RRT-based approach for planning a sequence of pushes. Similar to potential-field-based motion planners developed by \cite{khatib1986real}, \citet{igarashi2010pushdelivery} proposed a method that computes dipole-like vector fields around circular objects that guide the motion of a robot with a circular manipulator.

\begin{figure}[t]
    \centering{
        \includegraphics[scale=0.60]{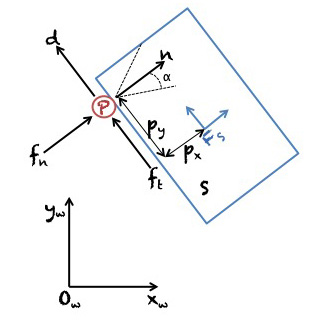}
        \includegraphics[scale=0.60]{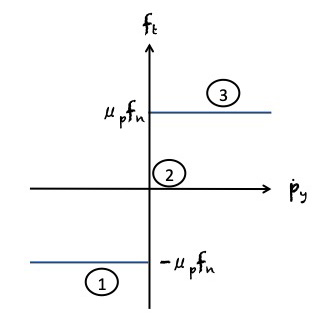}
    }
    \caption{Left. Planar pushing system with world frame $F_w$ (with origin $O_w$) and a slider $S$ (blue) with frame $F_w$ as described in Fig.~\ref{fig:planar_pushing}. The pusher $P$ (red) is interacting with the slider on one point of contact. It impresses a normal force $f_n$, a tangential friction force $f_t$, and a torque $\tau$ about the centre of mass. The normal force $f_n$ is in the direction of the normal vector $n$ of the contact point between pusher and slider, and $\alpha=tan^{-1}\mu_p$ is the angle of the friction cone assuming $\mu_p$ as the coefficient of friction. The terms $p_x$ and $p_y$ describe respectively the normal and the tangential distance between the pusher $P$ and the centre of friction of the slider $S$. Right. Coulomb's frictional law for the planar pushing system on the left-hand figure. Coulomb's law states that the normal and tangential forces are related by $f_t=\mu_pf_n$. Three contact modes are defined. 1. Sliding right in which friction acts as a force constraint; 2. Sticking in which friction acts as a kinematic constraint; and 3. Sliding left in which friction acts as a force constraint.
    Reproduced from~\cite{bauza2018}.}
    \label{fig:planar_pushing_coulomb}
\end{figure}

More recently, \citet{dogar2011framework} employed the	ellipsoidal approximation of the limit surface to plan robust push-grasp actions for dexterous hands and used them for rearrangement tasks in clutter. To use results for planar pushing, they assumed that objects do not topple easily. Furthermore, they assumed that the robot has access to 3D models of the objects involved. \cite{cosgun2011pushplancluttered} presented an algorithm for placing objects on cluttered table surfaces, thereby constructing a sequence of manipulation actions to create space for the object. However, focusing on planning, in their 2D manipulation they simply push objects at their centre of mass in the desired direction. \citet{lee2015multicontact} presented a hierarchical approach to planning sequences of non-prehensile and prehensile actions, proceeding in three stages. First, they find a sequence of qualitative contact states of the moving object with other objects, then a feasible sequence of poses for the object, and lastly a sequence of contact points for the manipulators on the object. \citet{king2016nonprehensile} developed a series of push planners for open-loop non-prehensile rearrangement tasks in cluttered environments. Before considering more complex scenarios, they used a simple analytical approach for forward-simulation of randomly sampled time-discrete controls within an RRT-based planner. They tested their planners on two real robotic platforms, the home care robot HERB with a seven DOF arm, and the NASA rover K-Rex.

\subsubsection{Complementing Analytical Approaches with Data-Driven Methods}

Transitioning to the second phase of planar pushing research, multiple factors have contributed a shift toward more \textit{data-driven approaches}. For one thing, much of the previous work makes minimal assumptions regarding the pressure distribution. While convenient, those methods lead to conservative strategies for planning and control, providing only worst case guarantees. Furthermore, while assumptions regarding the pressure distribution in previous work were often minimal, other strong assumptions were frequently made to derive results analytically. Hence, more recent work has set out to validate common assumptions such as the ubiquitous quasi-static assumption. Additionally, purely analytical models do not take into account the stochastic nature of pushing in the sense that pushes indistinguishable to sensor and actuator resolution have empirically been found to produce variable results \citep{yu2016data}. Instead of making minimal or strong assumptions about parameters, they can instead be \textit{estimated} based on observations. Several researchers have explored this approach.

\begin{figure}[t]
    \centering 
    \includegraphics[scale=0.6]{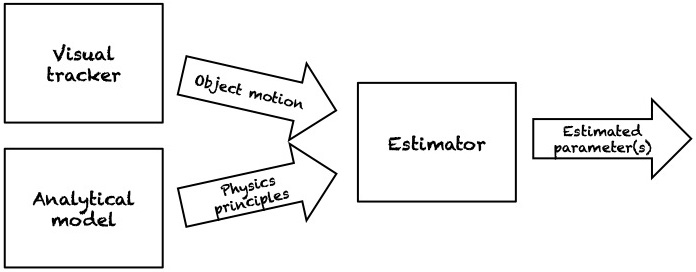}
    \caption{A classical workflow for estimating relevant physical parameters of a pushed object. A robotic pusher performs a set of push operation on an object which is typically tracked using vision. Simpler approach employs markers on the object for more accurate estimations. An analytical model of the motion for the target object is also employed. Sensory data and physical principles are the inputs of the estimator. As output, the estimator provides with an estimate of the desired parameters, e.g. friction distribution or centre of mass. In \citet{lynch1993estimate} the estimated parameters are also used for recognising objects based on their (estimated) physical properties.}
    \label{fig:planar_pushing}
\end{figure}

\citet{lynch1993estimate} presented  methods both for estimating the relevant friction parameters by performing experimental pushes, and for recognising objects based on their friction parameters.  Similarly, \citet{yoshikawa1991estimate} described how a mobile robot with a visual sensor can estimate the friction distribution of an object and the position of the centre of friction by pushing and observing the result. Yet, both of these approaches discretise the contact patch into grids so that they are either imprecise if the approximation is too coarse or suffer from the curse of dimensionality when using a fine-grained approximation. \citet{ruizugalde2010objectmodel, ruizugalde2010effectawarepushing} formulated a compact mathematical model of planar pushing. Assuming that the object's base shape is given, their robot  explored object-table and finger-object friction coefficient parameters. \citet{zhou2016polynomial} developed a method for modelling planar friction, proposing a framework for representing planar sliding force-motion models using convex polynomials. Notably, they also showed that the ellipsoid approximation is a less accurate special case of this representation. \citet{zhou2017contactmodel} extended the convex polynomial model to associate a commanded position-controlled end effector motion to the instantaneous resultant object motion. They modelled the  probabilistic nature of object-to-surface friction by sampling parameters from a set of distributions. They presented the motion equations for both single and multiple frictional contacts and validated their results with robotic pushing and grasping experiments on the dataset published by \citet{yu2016data}. That dataset comprises planar pushing interactions with more than a million samples of positions of pusher and slider, as well as interaction forces. Push interaction is varied along six dimensions, namely surface material, shape of the pushed object, contact position, pushing direction, pushing speed, and pushing acceleration. Using their dataset, they characterised the variability of friction, and evaluated the most common assumptions and simplifications made by previous models of frictional pushing. They provide an insightful table that lists  the assumptions and approximations made in much of the work cited in this section. Finally, \citet{bauza2017planarprob} used a data-driven approach to	model planar pushing interaction to predict both the most likely outcome of a push and, as a novelty, its expected variability. The learned models (also trained on the dataset by \citet{yu2016data}) rely on a variation of Gaussian processes whilst avoiding and evaluating the quasi-static assumption. However, the  learned models are specific to the particular object and material. Transfer learning is left for future work. 

\subsection{Physics Engines and Dynamic Analysis}
			
While the quasi-static assumption may be reasonable in a variety of situations, other problems call for dynamic models of pushing. One popular approach to achieving this is using a physics engine. Before covering this field, we first consider work concerned with dynamic  pushing  that does not recur to physics engines.

\subsubsection{Dynamic Analysis}

Using dynamic analysis, \citet{brost1992dynamic} investigated the problem of catching an object by pushing it, i.e. determining the pushing motions that lead to a pusher-object equilibrium. This work was motivated by dealing with uncertainty in positioning, generating plans that work also in the worst case. \citet{jia1999pose} investigated dynamic pushing  assuming frictionless interaction between pusher and object. \citet{behrens2013dynamic} instead studied dynamic pushing but assumed infinite friction between pusher and object. \cite{chavan2015inhand} considered planning non-prehensile in-hand manipulation with patch contacts. They described the quasi-dynamic motion of an object held by a set of frictional contacts when subject to forces exerted by the environment. Given a grasp configuration, gripping forces, and the location and motion of a pusher, they estimate both the instantaneous motion of the object and the minimum force required to push the object into the grasp. To this end, complex contact geometries are broken up into rigid networks of point contacts.
			
\subsubsection{Physics Engines}
		
A large body of work related to pushing makes use of \textit{physics engines}. Commonly used examples of such engines include Bullet Physics, the Dynamic Animation and Robotics Toolkit (DART), MuJoCo, the Open Dynamics Engine (ODE), NVIDIA PhysX, and Havok  \citep{erez2015physicsengines}. Those engines allow for 3D simulation but 2D physics engines exist, as well, e.g. Box2D. While some physics engines have been designed for graphics and animation, others have been developed specifically for robotics. In the first category, visually-plausible simulations are key while  physically-accurate simulations are essential for many robotics applications. Most physics engines today use impulse-based velocity-stepping methods to simulate contact dynamics. As this requires solving NP-hard problems at each simulation step, more tractable convex approximations have been developed, highlighting the trade-off between computational complexity and accuracy present in those engines \citep{erez2015physicsengines}. 3D physics engines use a Cartesian representation where each body has six DOF and joints are modelled as equality constraints in the joint configuration space of the bodies. In robotics, where joint constraints are ubiquitous, using generalised coordinates is computationally less expensive and prevents joint constraints from being violated.
            
            
For a \textit{comparison} of physics engines, we refer the reader to two recent studies \cite{erez2015physicsengines, chung2016physicsengines}. \citet{erez2015physicsengines} compared ODE, Bullet, PhysX, Havok, and MuJoCo. It should be noted that the study was written by the developers of MuJoCo. They introduced quantitative measures of simulation performance and focused their evaluation on challenges common in robotics. They concluded that each engine performs best on the type of system it was tailored to. For robotics, this is MuJoCo while gaming engines shine in gaming-related trials, whereby no engine emerges as a clear winner. \citet{chung2016physicsengines} compared Bullet, DART, MuJoCo, and ODE with regard to contact simulations whilst focusing on the predictability of behaviour. Their main result is that the surveyed engines are sensitive to small changes in initial conditions, emphasising that parameter tuning is important. Another evaluation of MuJoCo was carried out by \citet{kolbert2017contactdynamics} who evaluated the contact model of MuJoCo with regard to predicting the motions and forces involved in three in-hand robotic manipulation primitives, among them pushing. In the course, they also evaluated the contact model proposed by \cite{chavan2015inhand}. They found that both models make useful yet not highly accurate predictions. Concerning MuJoCo, they state that its soft constraints increase efficiency but limit accuracy, especially  in the cases of  rigid  contacts  and  transitions  in  sticking and slipping at contacts.

Researchers have  \textit{applied} physics engines in multifarious ways to study robotic pushing. To begin with, physics engines have been used in RRT-based planners to forward-simulate pushes. \citet{zito2012two} presented a two-level planner that combines a global RRT planner operating in the configuration space of the object, and a local planner that generates sequences of actions in the robot's joint space that will move the object between a pair of nodes in the RRT. In this work, the experimental set-up consists of a simulated model of a tabletop robot manipulator with a single rigid spherical fingertip which it uses to push a polyflap \cite{sloman2006} to a goal state. To achieve this, the randomized local planner utilizes a physics engine (PhysX) to predict the object's pose after a pushing action. Similarly, \citet{king2016nonprehensile} incorporated a dynamic physics engine (Box2D) into an RRT-based planner to model dynamic motions such as a ball rolling. To reduce planning complexity, they considered only dynamic actions that lead to statically stable states, i.e. all considered objects need to come to rest before the next action. Another application of physics engines in robotic pushing was proposed by \citet{scholz2014oomdp}. In what they refer to as  Physics-Based  Reinforcement Learning, an agent uses a physics engine  as a model  representation. Hence, a physics engine can be seen as a hypothesis space for  rigid-body dynamics. They introduced uncertainty using distributions over the engine's physical parameters and obtained transitions by taking the expectation of the simulator's output over those random variables. Finally, \citet{zhu2017physicsengine} utilised a physics engine for motion prediction,  learning the physical parameters through black-box Bayesian optimization. First, a robot performs random pushing actions on an object in a tabletop set-up. Based on those observations, the Bayesian learning algorithm tries to identify the model parameters that maximise the similarity between the simulated and observed outcomes. To support working with different objects, a pre-trained object detector is used that maps observed objects to a library of 3D meshes and  estimates the objects' poses on that basis. Once the physical parameters  have been identified, they are used to simulate the results of new actions.	
\begin{figure}[t]
  \centering 
  \includegraphics[scale=0.25]{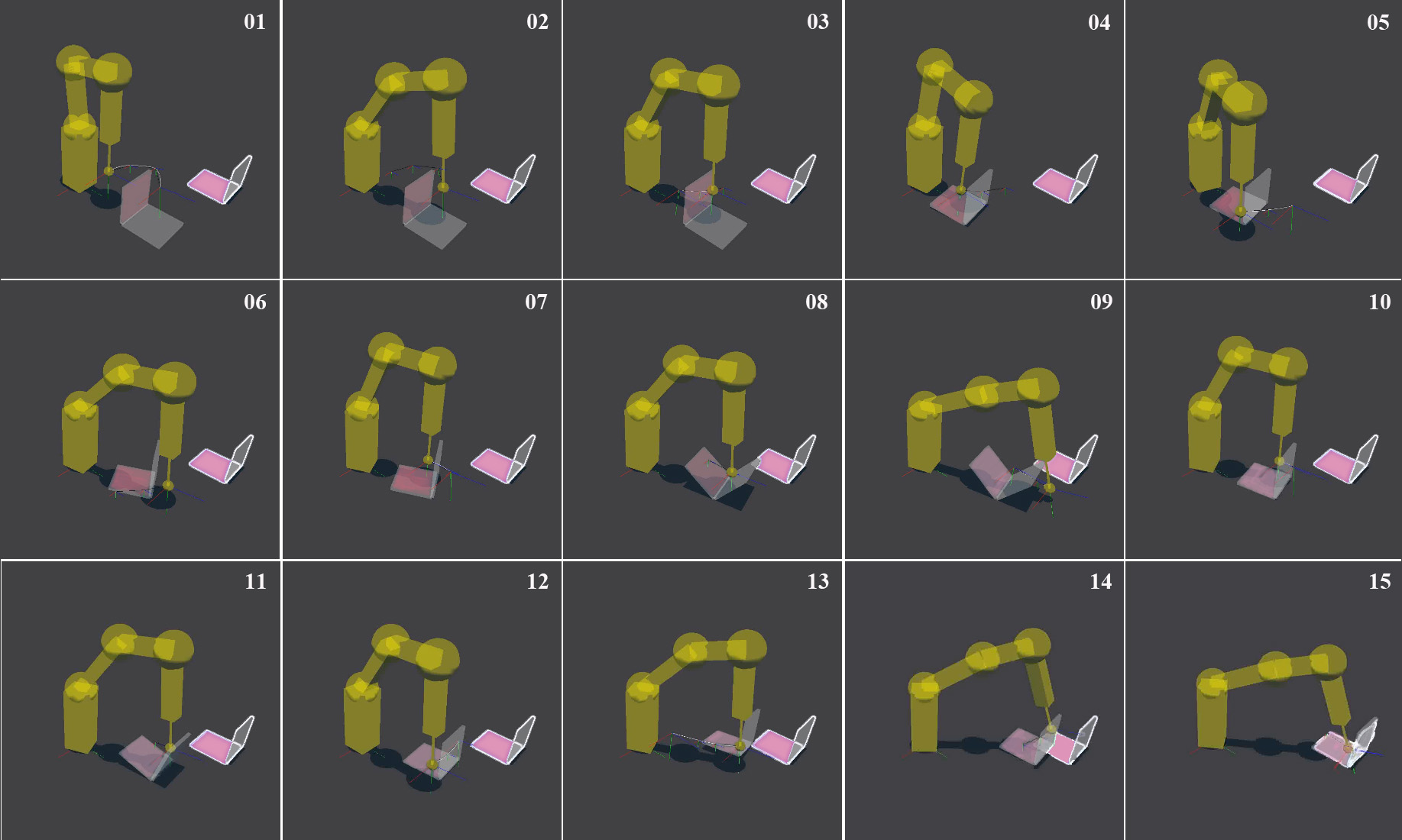}
  \caption{Simulation of a Katana robot arm equipped with a spherical finger that plans a sequence of pushes to move an L-shaped object, called polyflap~\cite{sloman2006} to a goal state. The plan is created by using a physics engine (PhysX) to predict the outcome of a push operation. Image 01 shows the initial pose. The wire-framed L-shaped polyflap is a `phantom' to indicate the desired goal state. The goal pose is translated from the initial pose by 28 cm and rotated by 90 degrees. Image 02 shows the collision-free trajectory to bring the end effector to the start pose of the first push. Images 01-04 show the first push which makes the polyflap tip over. Images 05-09 show a series of pushes which culminate in the polyflap resting in an unstable equilibrium pose along its folded edge. Images 12-13 show a sideways push. Images 14-15 show the final frontal push which aligns the polyflap with the target configuration. Courtesy of~\citet{zito2012two}.}
  \label{fig:strip1}
\end{figure}

While physics engines offer great value for robotic applications, e.g. by taking into consideration dynamic interaction and 3D objects, they nevertheless require explicit object modelling and extensive parameter tuning. Another approach, which we consider next, is to learn a forward model from data. 

\section{Learning to Predict from Examples}

This part of the literature is based on learning forward models for robotic pushing from data. We first review work on \textit{qualitative models} and then consider models that make \textit{metrically precise} predictions. In both of those sections, we do not include work that uses \textit{deep learning} techniques. We dedicate a separate section to such approaches, given the current research interest in that area and the large number of papers being published. 

\subsection{Qualitative Models}
        
Much work on qualitative models revolves around the concept of \textit{affordances}. The term affordance was invented by \citet{gibson1979affordance} and generally refers to an action possibility  that an object or environment provides to an organism. Although it has originated from psychology, the concept has been influential in various domains, among them robotics. \citet{sahin2007affordances} discussed affordances from a theoretical perspective while laying emphasis on their use in autonomous  robotics. \citet{min2016affordancesurvey}  provided a recent survey of affordance research in developmental robotics. \citet{ugur2011affordances} considered an anthropomorphic robot that learns object affordances from observations with a range camera and uses them for planning. First, the robot discovers effect categories from its actions. Then, it learns a mapping between object features and affordances which it then employs for planning.  Pushing is one type of actions that they consider. While much previous work  has focused  on affordance models  for  individual  objects, \citet{moldovan2012learning} learned  affordance  models for  configurations of multiple interacting objects.  Their model is capable of generalising  over  objects  and dealing with uncertainty. \citet{ridge2015pushaffordances} developed a self-supervised online learning framework based on vector quantization for acquiring models of effect classes and their associations with object features. Specifically, they considered robots pushing household objects and observing them with a camera. Limitations of such approaches are that they do not tend to generalise well to novel objects and actions.  

Considering \textit{other qualitative approaches} than those related to affordances, \citet{zrimec1991pushlearninduction} developed an algorithm for  knowledge   extraction and representation to predict the effects of pushing. In their experiment, a robot performs random pushes and uses unsupervised learning on those observations. Their method involves partitioning, constructive induction and determination of dependencies. \citet{hermans2013locations} developed a method for predicting contact locations for pushing based on the global and local object shape. In exploratory trials, a robot pushes different  objects, recording the objects' local and global shape features at the pushing contacts. For each observed trajectory, the robot computes a push-stability or rotate-push score and maps shape features to those scores by means of regression. Based on that mapping, the robot can search objects of novel shape for features associated with effective pushes. Experimental results are reported for a mobile manipulator robot pushing household objects in a tabletop set-up.

\begin{figure}[t]
  \centering 
  \includegraphics[scale=0.6]{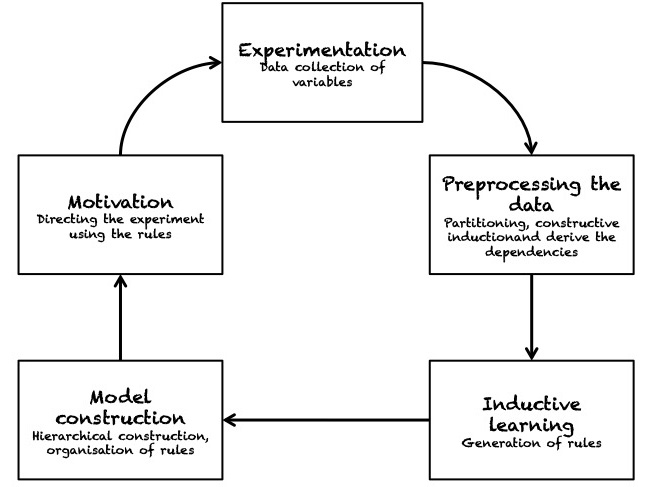}
  \caption{The sequence of operations adopted by~\citet{zrimec1991pushlearninduction} to construct their casuality learning model. The robot learns by interacting with the environment in an unsupervised fashion. The system can autonomously discover knowledge, as e.g. whether an action generates a push on an object. The ``motivation'' module guarantees that the system is driven towards acquiring more knowledge about the robot/environment interaction. Reproduced from~\citet{zrimec1991pushlearninduction}.}
  \label{fig:casuality}
\end{figure}

While learned affordances, and other qualitative models, can be useful in various scenarios, other applications require the ability to predict the effects of pushing more precisely, e.g. by explicitly predicting six DOF rigid body motions. We consider efforts made to achieve precise predictions in the next section.

\subsection{Metrically Precise Models}
        
Early seminal work by \citet{salganicoff93avisionbased} presented a vision-based unsupervised learning method for robot manipulation. A robot pushes an object at a rotational point contact and learns a forward model of the action effects in image space.  Subsequently, they used the forward model for stochastic action selection in manipulation planning and control. The scenarios considered in this work are relatively simple in that the pusher remains within the friction cone of the object and the contact only has one rotational DOF. Yet, this work takes an approach that is markedly different from analytical models discussed before. Instead of estimating parameters such as frictional coefficients explicitly, the authors \textit{encode} that information \textit{implicitly} in the mapping between actions and their effects in image space. Similarly, \citet{walker2009maps} learned a mapping between pushes and object motion as an alternative to explicitly modelling support friction. Set in a 2D  tabletop environment,  a  robot with a single  finger pushes  objects  and uses an online,  memory-based local regression model to learn manipulation maps. To achieve this, they explicitly detect the object's shape using a proximity sensor and  fit a shape to the thus obtained point cloud. A method for handling objects with more complex shapes was proposed by \citet{lau2011pushstrategy}. In their work, a robot, while being of simple  circular shape itself, aims to deliver irregularly shaped flat objects to a goal position by pushing them.  The objects that they consider are chosen to exhibit quasi-static properties. Collecting several hundred random example pushes as training data, a forward model is learned using non-parametric regression, similar to the approach taken by \citet{walker2009maps}. 

\begin{figure}[t]
  \centering
  \includegraphics[scale=0.8]{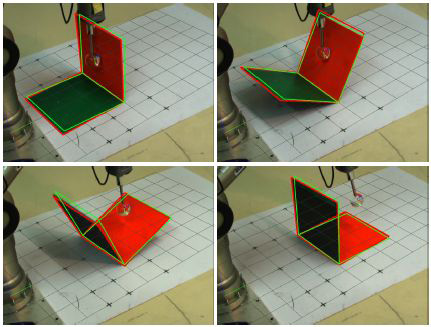}
  \caption{The example shows the interaction between a 5-axis Katana robotic manipulator and an L-shape object, called polyflap~\cite{sloman2006}. A set of contact experts are learned as probability densities for encoding geometric relations between parts of objects under a push operation. This approach allows these experts to learn from demonstration physical properties, such as non-penetration between an object and a table top, without explicitly representing physics knowledge in the model.    
  The green wire frame denotes the prediction whilst the red wire frame denotes the visual tracking. Courtesy of~\citep{kopicki2011predictrigid}.}
  \label{fig:marek}
\end{figure}

\citet{kopicki2011predictrigid} presented two data-driven probabilistic for predicting 3D motion of rigid bodies interacting under the quasi-static assumption. First, they formulated the problem as regression and subsequently as density estimation. In \cite{kopicki2016forwardmodel} they extended this work further. 
Their architecture is modular in that multiple object- and context-specific forward models are learned which represent different constraints on the object's motion. A product of experts is used which, contrary to mixture models, does not add but multiply different densities. Hence, all constraints, e.g. those imposed by the robot-object contact and multiple object-environment contacts, need to be satisfied so that a resulting object motion is considered probable. This formulation facilitates the transfer of learned motion models to objects of novel shape and to novel actions. In experiments with a robot arm, the method is compared with and found to outperform the physics engine PhysX tuned on the same data. For learning and prediction, their algorithms require access to a point cloud of the object. 
A further extension of this approach is presented in~\citet{stuber2018icra}. In this work, the authors aim to contribute to endowing robots with versatile non-prehensile manipulation skills. To that end, an efficient data-driven approach to transfer learning for robotic push manipulation is proposed. This approach combines and extends two separate strings of research, one directly concerning pushing manipulation \citep{kopicki2016forwardmodel}, and one originating from grasping research \citep{kopicki2016oneshot}. The key idea is to learn motion models for robotic pushing that encode knowledge specific to a given type of contact (see the work in \citep{kopicki2016oneshot} for further details). In an previously unseen situation, when the robot needs to push a novel object, the system first establishes how to create a contact with the object's surface. Such a contact is selected among the learned models, e.g. a flat contact with a cube side or a contact with a cylindrical surface. At the generated contact, the system then applies the appropriate motion model for prediction, similarly to \citep{kopicki2016forwardmodel}.
The underlying rationale for this approach to prediction is that predicting on familiar ground reduces the motion models' sample complexity while using local contact information for prediction increases their transferability.  

\citet{mericli2015complexpush} similarly presented a case-based approach to push-manipulation prediction and planning. Based on experience from self-exploration or demonstration, a robot learns multiple discrete probabilistic motion models for pushing complex 3D objects on caster wheels with a mobile base in cluttered environments. Subsequently, the case models are used for planning pushes to navigate an object to a goal state whilst potentially pushing  movable obstacles out of the way. In the process, the robot continues to observe the results of its actions and feeds that data back into the case models, allowing them to improve and adapt.

\begin{figure}[t]
\centering{
\includegraphics[width=0.7\columnwidth]{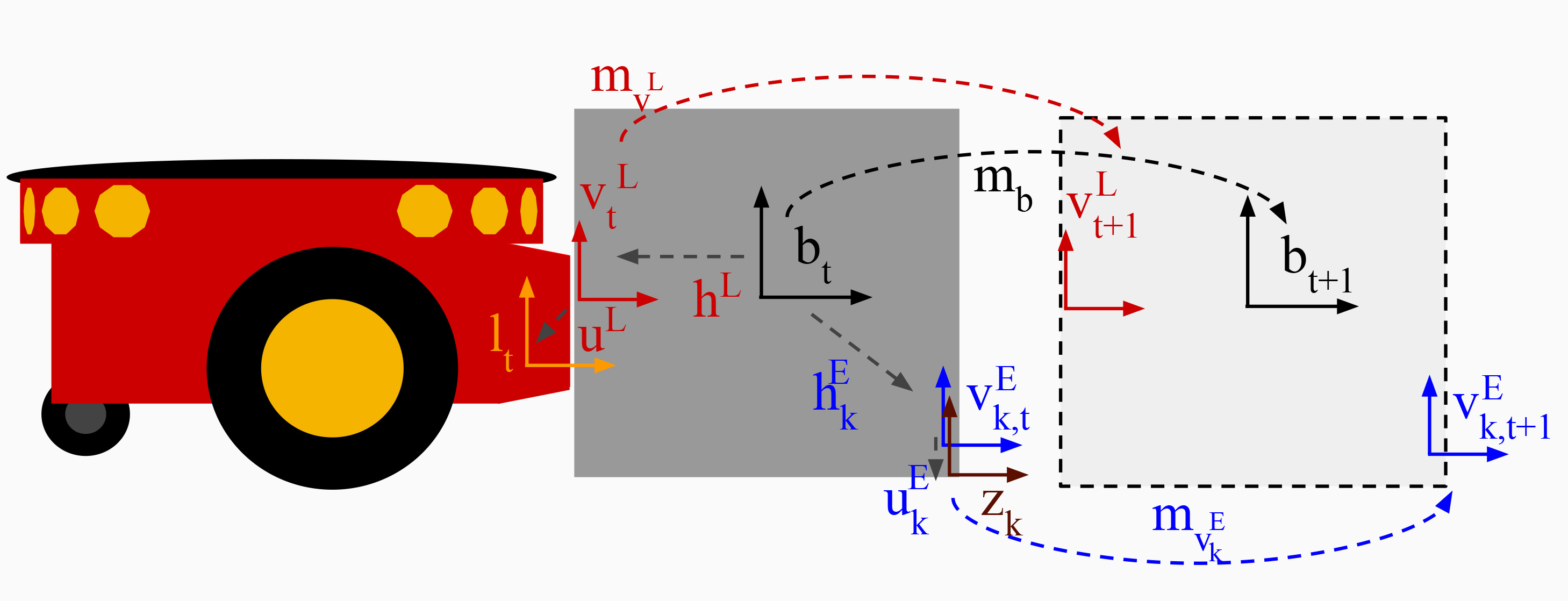}
\includegraphics[width=0.8\columnwidth]{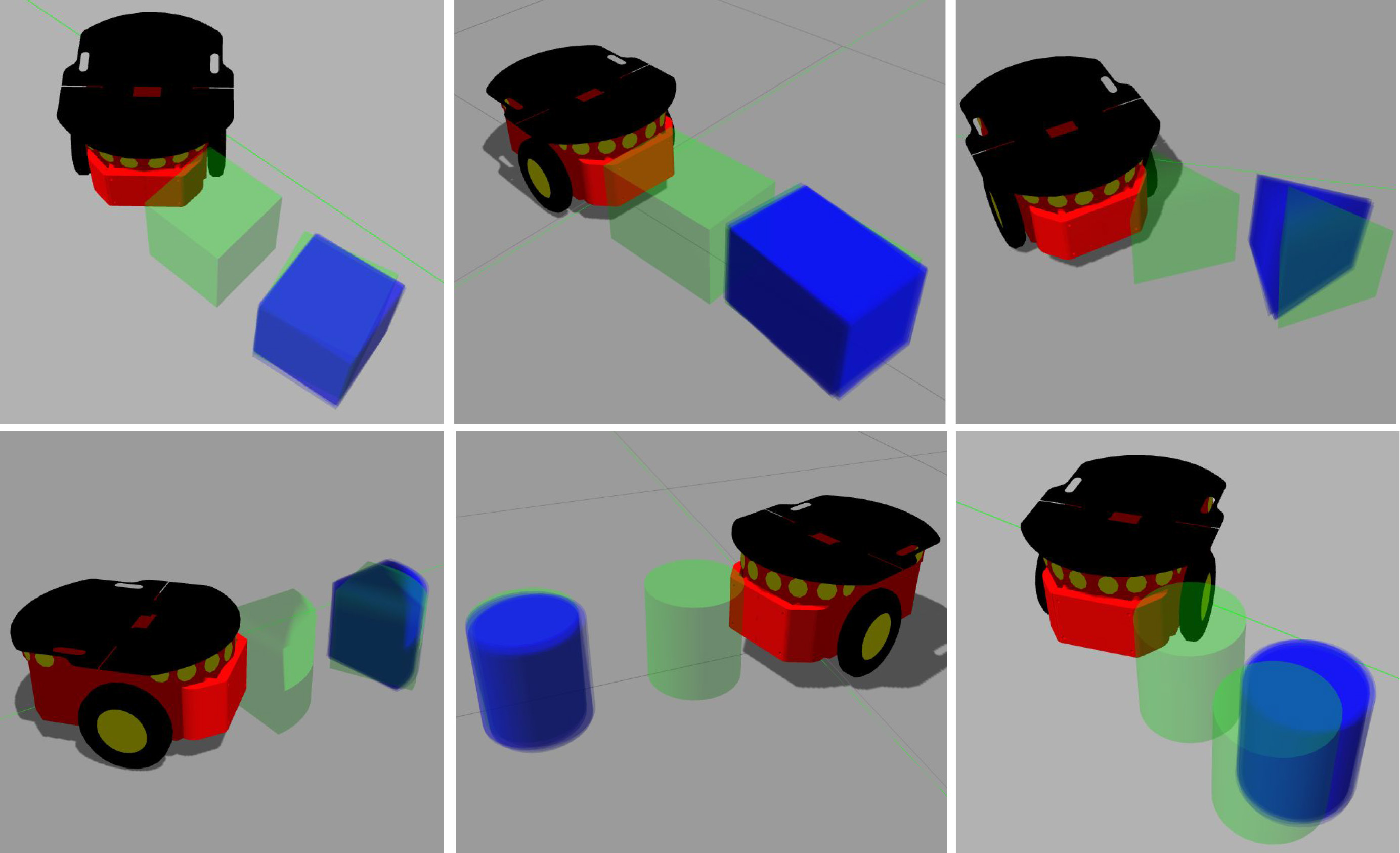}
}
\caption{Top: graphical representation of the feature-based predictors for push operations. The global motion of the object after a push is described by the rigid body transformation $m_b$. This transformation is unknown to the robot. However, the robot can estimate it by learning a set of local predictors for the motions $m_{v^L}$ and $m_{v_k^E}$, for $k=1,\dots,N_E$. The rigid body transformations $h^L$ and $h^E_k$ describe the estimated contacts on the object's surface w.r.t. the estimated global frame of the object, $b$. Since the object is assumed to be rigid, this relation does not change over time, thus once the local motions $m_{v^L}$ and $m_{v_k^E}$ are estimated, $b_{t+1}$ can be estimated by using the relations $h^L$ and $h^E_k$. 
Bottom: resulting predictions. initial object pose (green, in contact with robot), true final object pose (green, displaced), and predictions (blue).
Courtesy of~\citet{stuber2018icra}.
}
\label{fig:frames}
\end{figure}

\subsection{Deep Learning Approaches}

\textit{Artificial neural networks} have been used in robotic pushing to estimate physical parameters, predict the outcome of pushing actions, and for planning and control. Previously, we have seen work concerned with \textit{estimating physical parameters} of the environment from data. Deep learning has been used to address the same problem. \citet{denil2016physicsexperiments} studied the learning of physical properties such as mass and cohesion of objects in a simulated environment. Using deep reinforcement learning, their robots learn different strategies that balance the cost of gathering information against the cost of inaccurate estimation.

Instead of explicitly estimating physical parameters, another approach is learning a \textit{dynamics model}. Several studies have investigated learning general physical models or "physical intuition" directly from image data. \citet{chang2016deepdynamics} presented the Neural Physics Engine, a deep learning framework for learning simple physics simulators. They factorise the environment into object-based representations and decompose dynamics into pairwise interaction between objects. However, their evaluation is limited to simple rigid body dynamics in 2D. 

\citet{watters2017interactnetworks}  introduced the Visual Interaction Network, a model for learning the dynamics of a physical system from raw visual observations. First, a convolutional neural network (CNN) generates a factored object representation from visual input. Then, a dynamics predictor based on interaction networks computes predicted trajectories of arbitrary length. They report accurate predictions of trajectories for several hundred time steps using only  six input video frames.  Yet, their experiments are also limited to rather simple environments, namely 2D simulations of coloured objects on natural-image backgrounds.  Similarly, \citet{fragkiadaki2015predictbilliard} also used an object-centric formulation based on raw visual input for dynamics prediction. Based on object-centric visual glimpses (snippets of an image), the system predicts future states by individually modelling the behaviour of each object. After training in different environments by means of random interaction, they also use their model for planning actions in novel environments.  Again, they consider simple 2D worlds, in this case moving balls on a 2D table, i.e. their agent plays 2D billiards. \citet{ehrhardt2017deepphysics}  constructed a neural network for end-to-end prediction of mechanical phenomena. Their architecture consists of three components: a CNN extracts features from images which are updated by a propagation module,  and decoded by an estimation  module. What their network outputs is a distribution over outcomes, thus explicitly modelling the inherent uncertainty in manipulation prediction. In terms of experiments, they study the relatively simple problem of a small object sliding down an inclined plane. 

\begin{figure}[t]
\centering
\includegraphics[width=0.7\columnwidth]{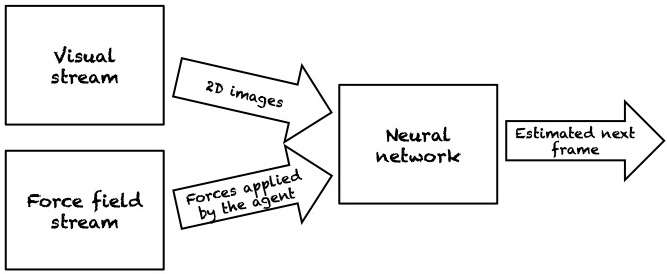}
\caption{Frame-centric model for motion prediction of billiard balls. The model takes as input the 2D image of the billiard and the forces applied by the agent to make predictions about the future configurations of the balls.
Reproduced from~\citet{fragkiadaki2015predictbilliard}.
}
\label{fig:frame_centric}
\end{figure}

Moving towards more complex scenarios, \citet{byravan2016deeprigid} introduced SE3-NETS, a deep neural network architecture for predicting 3D rigid body motions. Instead of RGB images, their network takes depth images as input, together with continuous action vectors, and associations between points in subsequent images. SE3-NETS segment point clouds into object parts and predict their motion in the form of $SE(3)$ transformations. They report that their method outperforms flow-based networks on simulated depth data of a tabletop manipulation scenario. Furthermore, they demonstrate that it performs well on real depth images of a Baxter robot pushing objects. However, their approach requires that associations between depth points are provided. They aim to learn those automatically in future work and to apply SE3-NETS to non-rigid body motion, recurrent prediction, and control tasks. A different approach to learning dynamics from images was taken by \citet{agrawal2016pokelearn}. They jointly learn forward and inverse models of dynamics of robotic arm operation that can be used for poking objects. In doing so, they extract features from raw images and make predictions in that feature space. In real-world experiments with Baxter, their model is used to move objects to target locations by poking. In order to cope with the real world, their model requires training on large amounts of data. By poking different objects for over $400$ hours, their robot observed more than $100,000$ actions. 

\begin{figure}[t]
\centering
\includegraphics[width=0.7\columnwidth]{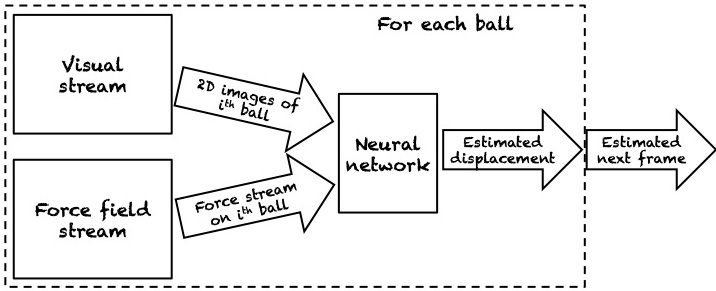}
\caption{Object-centric model for motion prediction of billiard balls. The system predicts the future configurations of the balls by individually modelling the temporal evolution of each ball. In this scenario, predicting the velocities of each ball is sufficient for computing the next configuration of the billiard. 
Reproduced from~\citet{fragkiadaki2015predictbilliard}.
}
\label{fig:object_centric}
\end{figure}
Most of the studies presented this section make use of \textit{object-centric representations} to model dynamics. Approaches that predict motion without such a representation have been explored, as well. For instance, \cite{finn2016unsupervisedphysics} developed an action-conditioned video prediction system which predicts a distribution over pixel motions only based on previous frames. No information concerning object appearance is provided to the model. It borrows that information from previous frames and merges it with model predictions. It is this mechanism that allows the model to generalise to previously unseen objects. By conditioning predictions on an action, the model can effectively imagine the action's consequences. As with previously presented deep learning models, this approach also requires large amounts of data to perform well in real-world situations. Hence, the authors have collected a dataset of $59,000$ robot pushing motions (frames associated with the action being applied) on different objects. While their results demonstrate that no object-centric representation is required for prediction, the authors argue that such representations are a promising direction for research as they provide concise state representations for use in reinforcement learning. 

We have seen how artificial neural networks can be used to model the dynamics of physical systems. In addition to that, deep reinforcement learning has been used to learn control policies in the field of robotic pushing. Many of those approaches make use of dynamics models so that they can be seen as complementary to the work presented before.    We do not provide a detailed review of this very active field here, rather referring the reader to \cite{levine2015contactrich}, \citet{levine2016deependtoend}, \citet{finn2017deepforesight},  and \citet{ghadirzadeh2017deepcontrol} for overviews of such work.

\section{Final Remarks}

In this paper we have provided an overview of the problem of robot pushing and summarised the development of the state-of-the-art, focusing on the problem of motion prediction of the object to be pushed. We have also covered some aspects of relevant applications of pushing for planning and control. 

Typical approaches have been classified as i) purely analytical, ii) hybrid, iii) dynamic analysis, iv) physics engines based, v) data-driven, and vi) deep learning. Representative work for each of these categories has been listed for readers to have a general overview of the field and its state-of-the-art from the earlier work in the 1980s to the most recent approaches. 

A set of assumptions in the proposed methods have been highlighted. Earlier work has mostly investigated motion prediction with the quasi-static assumption to get rid of complex dynamics and provided the groundwork to understand the mechanics for pushing 2D shapes~\citet{mason1986quasistatic}. This seminal work has been extensively extended to more realistic scenarios involving 3D object to be pushed.
Nonetheless, as we have seen there are two types of uncertainty that affect manipulation problems: i) prediction uncertainty and ii) state uncertainty. The majorities of the papers that investigated the extension to 3D object commonly relied on the assumption that the geometrical properties of the object to be pushed were known \textit{a priori}, e.g.\citet{mason1990blockwall,mayeda1991wallpush}. Key physical properties that would affect the prediction, e.g mass distribution or friction coefficients, were typically assumed to be known or possible to estimate on the fly, as in~\citet{yoshikawa1991estimate}, by combining data-driven methods to the analytical mechanics of pushing.  

More recently, a few efforts were made towards robot pushers that can also deal with state uncertainty. By relaxing the assumption that the model of the object to be pushed is known, the robot typically perceives the object as a point cloud or RGB image to estimate the geometric properties, such as pose and shape, before even attempt to make a motion prediction, see~\citet{fragkiadaki2015predictbilliard,stuber2018icra}. 

Two strands of approaches can be identified. First, the data-driven approach that attempts to learn from experience how an object behaves under a push operation. Qualitative models have investigated the concept of affordances for learning a mapping between object features and possible push actions, which they then employ for planning, e.g.~\citet{zrimec1991pushlearninduction}. In contrast, metrically precise models have investigated how to learn a mapping between actions and its effects, e.g.~\citep{kopicki2016forwardmodel}. A second more recent strand is the application of deep learning techniques to learn a physical intuition of the mechanics of pushing from visual data, see~\citet{fragkiadaki2015predictbilliard}. Both strands typically model the predictions in a probability framework to estimate the most likely action's outcome given the information available, e.g. image of the scene, contact models. The latter approach is very promising, but it requires a massive amount of data for the model to learn. Hence, such approaches are typically relying on synthetic data from physics engines. In contrast, the work by~\citet{stuber2018icra,mericli2015complexpush} have demonstrated that it is possible to learn motion prediction for complex push operations efficiently and generalise the model's predictions to previously unseen objects.     

While some typical problems still require a better solution, new challenges and requirements are emerging in the field. To make pushing an essential motor primitive in practical robotics, the challenges are either currently under investigation in research group worldwide or need to be investigated in the future. Following we list some suggested trends of open problems that we have identified.

\subsection{Understanding and Semantic Representation}

The scene is typically perceived as an RGB image or a point cloud. However, for robot pushing, we need to be able to identify pushable objects from static ones. Labelling can be done but it is very expensive in terms of human labour. Converting from source image data to geometrical shapes, and from geometrical shape to semantic representation will be beneficial for the robot. Once the robot can identify probable dynamic objects it would be able to interact with the environment prioritising those objects and improving its understanding.

\subsection{Sensory Fusion and Feedback}

Multiple sensor inputs are nowadays available for robotic system. Instead of solely relying on vision, other sources of information should be used to close the loop of the manipulation. Tactile, proprioception, and visual feedback should be fused together to enable the robot to perform complex manipulation and recover from failures.

\subsection{Explicitly Modelling Uncertainty in the Model}

Due to a lack of perfect perception abilities, it is not unusual that robot has to operate with an incomplete description of their environment. In robot pushing, but more in general in the problem of manipulation, the robot needs to generate a set of contacts to interact with other objects. When the pose of the object to be manipulated is unknown, what is the best way to create a robust set of contacts?
In the case of planning for dexterous manipulation, our previous work in~\citet{zito2013iros} has demonstrated that approaching directions that maximise the likelihood of gathering (tactile) information are more likely to achieve a successful set of contacts for a grasp. This was tested in the case when due to imperfect perception abilities the description of the object to be grasped results incomplete and hence the pose of the object is uncertain. This empirically suggests that reasoning about the uncertainty leads to more robust reach-to-grasp trajectories with respect to object-pose uncertainty. Similarly, selecting an action for physical effects (e.g. pushing, push and grasp) should benefit from incorporating state uncertainty with respect to the initial pose estimate of the object. 

\subsection{Cooperative Robots and Multiple Contacts Pushing}

In warehouses, for example, exists the problem of moving large-scale objects. Collaborative robots may be able to complete the task. Besides the problem of sharing sensitive information between them and coordinate the efforts, a new challenge arise from the manipulation point of view. Multiple contacts pushing is hard to predict, especially when the actions are carried by multiple agents. Control and decision making is a critical issue in such systems.

\subsection{Real-world Applications}

Although the theory behind motion prediction is well-established and applications to simple, structured scenarios have been made, the combination of the existing methods with any industrial applications has not been achieved yet. Robots in warehouses can navigate freely and deliver goods, however, no robotic system is capable of exploiting pushing operations to perform tasks such as inserting a box onto an over-the-head shelf. Theoretical solutions are rarely reliable in practical engineering applications, hence many sophisticated practical approaches will be needed in the future.


\section*{Author Contributions}

JS is the main author of this paper and collected the literature. CZ is the leading supervisor of this work and he has co-written the paper. RS has co-supervised and funded this project. 

\section*{Funding}
This work was supported by UK Engineering and Physical Sciences Research Council (EPSRC No. EP/R02572X/1) for the National Centre for Nuclear Robotics (NCNR).




\bibliographystyle{frontiersinSCNS_ENG_HUMS} 
\bibliography{source}

\end{document}